\documentclass[10pt,twocolumn,letterpaper]{article}

\usepackage{wacv}
\usepackage{times}
\usepackage{epsfig}
\usepackage{graphicx}
\usepackage{amsmath}
\usepackage{amssymb}
\usepackage{listings}
\usepackage{microtype}
\usepackage{subfigure}
\usepackage{amsfonts}       
\usepackage{nicefrac}       
\usepackage{amsthm}

\usepackage[retainorgcmds]{IEEEtrantools}
\usepackage{comment}
\usepackage{underscore}
\usepackage[english]{babel}
\usepackage{algorithm2e}
\graphicspath{{wacv2020-author-kit/latex/images/}}

\wacvfinalcopy 

\def\wacvPaperID{****} 

\ifwacvfinal\fi
\begin{document}

\title{An End-to-End Solution for Effectively Demoting Watermarked Images in Image Search}

\author{Ning Ma \hspace{0.2 cm} Xin Zhao\thanks{Currently a software engineer at Pinterest} \hspace{0.2cm} Mark Bolin\\
Microsoft\\
{\tt\small \{ninm, xinzhao, markbo\}@microsoft.com}
}

\maketitle
\ifwacvfinal\thispagestyle{empty}\fi

\begin{abstract}
We propose an end-to-end solution, from watermark feature generation to metric design, for effectively demoting watermarked images surfed by a real world image search engine. We use a few fundamental techniques to obtain effective watermark features of images in the image search index, and utilize the signals in a commercial search engine to improve the image search quality. We collect a diverse and large set (about 1M) of images with human labels indicating whether the image contains visible watermark. We train a few deep convolutional neural networks to extract watermark information from the raw images. The deep CNN classifiers we trained can achieve high accuracy on the watermark test data set. We also analyze the images based on their domains to get watermark information from a domain-based watermark classifier. We design a new novel hybrid metric which includes the relevance, image attractiveness and watermark information all together. We demonstrate that using these watermark signals together with the new metric in image search ranker can significantly demote the watermarked images during the online image ranking.
\end{abstract}

\section{Introduction}
\label{sec:intro}
\begin{figure}
\begin{center}
\includegraphics[width=1\columnwidth]{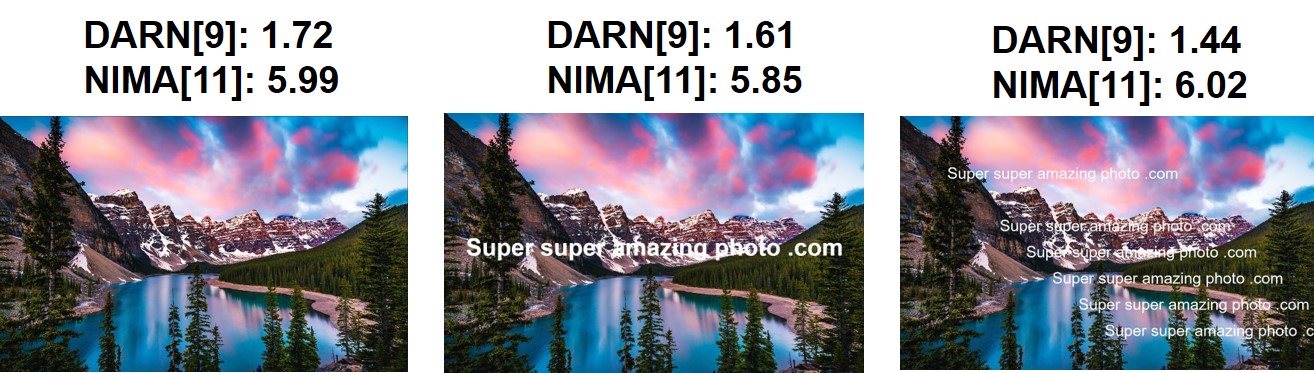}
\end{center}
   \caption{Model predicted image attractiveness score generally decreases as the image contains more visible watermarks. However, in the DARN\cite{attractivenessning} model, the score decreases very slowly as more and more watermarks added on the image. The NIMA\cite{Talebi2017} model has similar issue and even predicts that the image with most watermarks has highest score.}
\label{fig:attscore}
\end{figure}

Watermarking is a widely used technique to protect the copyright of image photography. There are a huge amount of watermarked images existing online. For example, a few famous image stock websites use watermarks to protect their high quality images from being copied by a third party. The drawback is that images with visible watermarks are often seen when customers are searching images on search engine like Bing, Google or Yahoo. The watermarked images can be annoying and degenerate customers' experience. Some researchers have looking into watermark removal \cite{WatermarkremovalGoogle, removal_Dashti, removal_Wang, removal_Yan} techniques to remove the watermark from the images or video. Most algorithms only work well in special situations, such as the \cite{WatermarkremovalGoogle} where the watermark has consistent pattern and does not have large variation. However, the watermark removal techniques are not very useful in image search mainly due to two reasons. First, the search engine should not remove the watermark before returning the search results to the users. This will remove the copyright protection for the original images and cause legal issues. Secondly, these techniques will not work well when the watermark has large variation which is exactly the case in a real image search index. Previous work \cite{attractivenessning, Talebi2017} demonstrates that an universal image attractiveness model can indicate the impact of the watermark by producing lower attractiveness score for the watermarked images than the original images. However, in the DARN model \cite{attractivenessning}, when the model predicts the score, it will take all possible image attributes into consideration, resulting in insignificant impact of watermark as shown in \textbf{Figure \ref{fig:attscore}}. It shows that the score will be significantly decreased only when there are massive watermarks on the original image. The NIMA\cite{Talebi2017} model even rates the image having most watermarks with highest score. It is likely due to the fact that the images in the AVA database does not contain watermarked images. So, purely using model based image attractiveness score to indicate watermark is not suitable as well. 

\begin{figure*}
\begin{center}
\includegraphics[width=1.8\columnwidth]{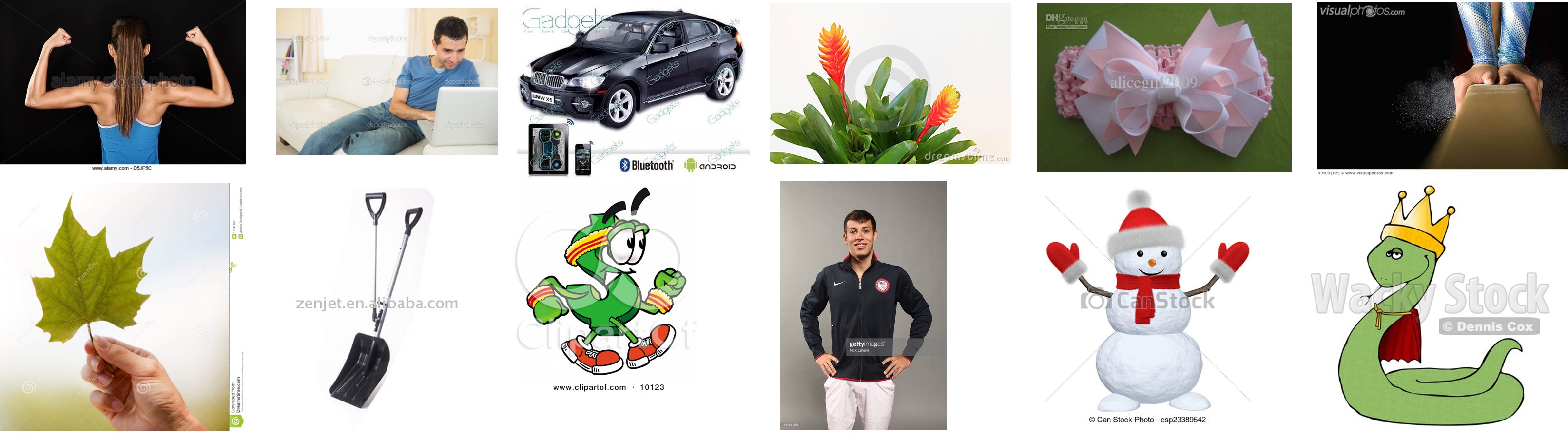}
\end{center}
   \caption{A few examples of watermarked images in this data set.}
\label{fig:watermark}
\end{figure*}

Regarding the application of image retrieval, a more appropriate approach is to demote images whose quality are significantly impaired by watermarks. In this paper, we propose a few fundamental techniques to obtain effective watermark signals for images coming from a real image search index, and utilize those watermark signals in a commercial search engine to improve the image search quality. Benefiting from the fast advance of deep learning, deep convolutional neural networks (CNN) have been widely used in image classification and detection tasks, and have achieved performance comparable to human. In section \ref{subsec:watermark_signal_content}, we train a few deep CNN models to predict the probability that an image contains a watermark using Resnet \cite{Resnet}, Densent \cite{Densenet}, and Inception-V3 \cite{InceptionV3} as the backbone. The model is trained end to end on a large image set with a variety of watermarks collected from real online images. The detail of the data set is described in section \ref{subsec:watermark_signal_content}. We show that the prediction accuracy of the deep CNN models are very promising on the data set with such diverse watermark patterns. This indicates the potential of building a DNN based universal watermark classifier. In section \ref{subsec:domain}, we also obtain an additional watermark signal by analyzing their corresponding domain properties. Our analysis indicates that domain is a very strong indicator of the watermark signal. This makes sense as a lot of watermarked images come from stock image website. However, in order to make these watermark features take effects in image ranker, the image ranker must have proper metric to reflect watermark information. In image retrieval, the metric used is the normalized discounted cumulative gain (NDCG) computed by $N_i = n_i\sum_{j=1}^{T}(2^{r(j)} - 1)/log(1 + j)$ where $r(j) \in \{0, ..., 4\}$ is the integer label for the relevance labels of $j^{th}$ URL in the sorted list and $n_i$ is the normalization factor. Since the rating score $2^{r(j)} - 1$ only considers the relevance, the image ranker will not pick up the watermark information even if we have watermark features available. In section \ref{sec:metric}, we introduce a novel hybrid metric which includes relevance, image attractiveness and watermark information in one place. We learn the weights between those factors from a side-by-side labeled data.  In section \ref{sec:onlinerank}, we demonstrate the effectiveness of demoting watermarked images in image search engine by utilizing those watermark signal and the new metric in the image ranker.

\section{Watermark Signal}
\label{sec:watermark_signal}
In this section, we demonstrate how we obtain watermark signals from two different approaches. The first approach is to get a watermark signal from the raw image content. This is a more biologically plausible method as humans only need to look at the raw image to tell if it contains a watermark. The second approach is from its corresponding domain information.
\subsection{Image content based watermark signal}
\label{subsec:watermark_signal_content}
A human can tell whether an image contains a watermark by directly looking at the image. Ideally, we should be able to train a similar classifier reflecting the probability that the image contains a watermark. The probability should reflect the visibility of the watermark in the images. Less visible watermarks should get lower probability.

\textbf{Data collection:}
We scraped a large amount of images from web image search results. For each image, we had 1-5 judges rate if this image contained a visible watermark. If any judge thought the image contained a visible watermark, the image would be labeled as positive, otherwise negative. Since the non-watermarked images are more than the watermarked images, we then randomly sample images from non-watermarked images, so that the watermarked and non-watermarked images are balanced. Next, we split the data into training, validation and test set with the rate 90\%:5\%;5\%. We also remove images which are broken or can not be downloaded. \textbf{Table~\ref{table:data}} shows the numbers of images we used to train and test the model. \textbf{Figure~\ref{fig:watermark}} shows a few examples of the watermarked images.

\begin{table}
\begin{center}
\begin{tabular}{|l|c|c|c|}
\hline
label & Traing & Validation & Testing\\
\hline\hline
1:Watermarked & 587K & 32K & 33K\\
\hline
0:No Watermark  & 646K & 36K & 36K\\
\hline
\end{tabular}
\end{center}
\caption{Summary of the watermark data set}
\label{table:data}
\end{table}

\begin{figure*}
\begin{center}
\includegraphics[width=2\columnwidth]{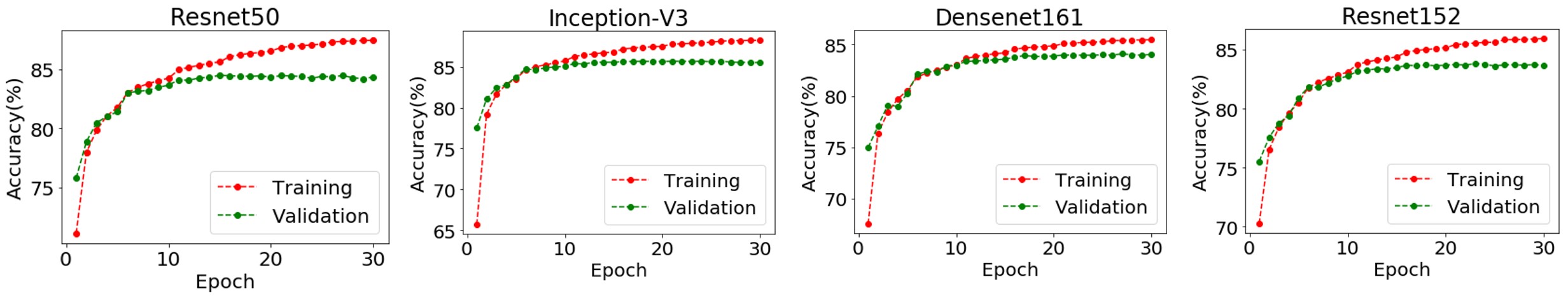}
\end{center}
   \caption{Training and validation accuracy progress of the four models during end-to-end training.}
\label{fig:finetuning}
\end{figure*}

\textbf{Data augmentation:}
We have about one millions images half of which have watermarks. During training, we did the following data augmentations to improve the performance. We used center cropping to obtain the images satisfying the input dimensional requirement of the different deep CNN models. Before cropping, we scaled the image dimension slightly larger than the model input dimension. We also used horizontal/vertical flips to increase the training dataset without losing the original watermark.

\textbf{Model:} We explored a few deep convolutioal neural network structures - Resnet50, Resnet152\cite{Resnet}, Densenet161\cite{Densenet}, and Inception\cite{InceptionV3}. We replace the final output classification layer with a binary classification layer. In the Inception-V3 model, we also replaced the intermediate auxiliary classification layer with a binary classification layer. The final loss function is $loss = loss_{out} + 0.4 * loss_{aux}$, where $loss_{out}$ is the cross-entropy loss function of the final output layer and the $loss_{aux}$ is the loss of the auxiliary classification layer.

\textbf{Training:} First, we use the transfer learning by freezing the models pretrained on ImageNet~\cite{ImageNet}, and only retraining the top and the auxiliary classification layer. The training error and validation error stops decreasing before ten epochs. \textbf{Table~\ref{table:transfer}} shows the accuracy of the models on the test data set after training 10 epochs. The ResNet152 obtained the best accuracy on the test data with 70.63\% accuracy. However, the overall accuracy of the transfer learning is low.  Next, we start training the whole network from end to end. \textbf{Figure~\ref{fig:finetuning}} shows the progress of the training and validation accuracy over epochs. We choose the model which performs best on the validation set and evaluate on the test set. The Inception-V3 has the best accuracy on the test set with 85.70\% accuracy. Both the validation and training accuracy are significantly improved after training the network end to end. This is likely because the the high level DNN features needed for watermark detection differ from general image classificaton. 

\begin{table}
\begin{center}
\begin{tabular}{|l|c|}
\hline
Model & Test Accuracy\\
\hline\hline
Resnet50 & 69.92\%\\
\hline
Inception-V3  & 64.12\%\\
\hline
Densenet161  & 68.93\%\\
\hline
Resnet152  & 70.63\%\\
\hline
\end{tabular}
\end{center}
\caption{Watermark prediction accuracy on the test data set, when retraining only the top classification layers}
\label{table:transfer}
\end{table}

During training, we set the learning rate as $1*10^{-4}$ and reduces it by half every 5 epochs. Unlike traditional fine tuning where the learning rate is set to be much smaller, we use the same learning rate and annealing procedure in both transfer learning and end-to-end training. This gives the model more freedom to discover the subtle watermark information. \textbf{Table \ref{table:finetuning}} shows the accuracy of the end-to-end retrained models on the test data set. Only the models with best performance on the validation set are evaluated on the test data. Our results show that the deep CNN can caputre the watermark signal from image pretty well. Also, training end-to-end significantly outperformed training just the final layers.

\begin{figure*}
\begin{center}
\includegraphics[width=2\columnwidth]{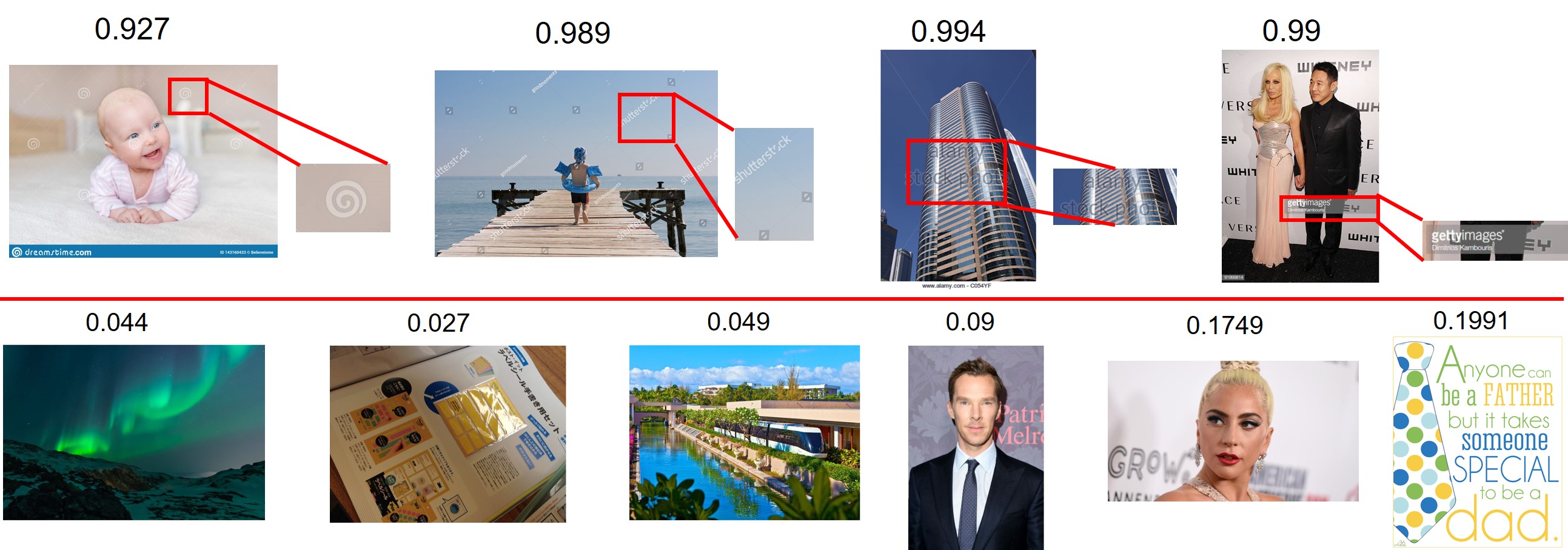}
\end{center}
   \caption{The predicted probability that the images contain visible watermark. The top row includes sample images detected including watermark. The bottom row shows the images detected without watermark.}
\label{fig:predwm}
\end{figure*}

Figure~\ref{fig:predwm} shows the prediction results using the trained resnet50 model. The images in the top row are the one detected with high probability of having a watermark. The bottom row includes the images detected with low probability of including watermark. We can see that the prediction is quite good. Another interesting thing we can observe is that the watermark is not simply just detecting text on the images. For example, the third, fourth, fifth and sixth images in the bottom row all contain texts, and all of them are successfully recognized as not containing watermark.

\begin{table}
\begin{center}
\begin{tabular}{|l|c|}
\hline
Model & Test Accuracy\\
\hline\hline
Resnet50 & 84.45\%\\
\hline
Inception-V3  & 85.70\%\\
\hline
Densenet161  & 83.96\%\\
\hline
Resnet152  & 83.86\%\\
\hline
Resnet50 + Domain & 87.04\%\\
\hline
Inception-V3 + Domain  & 87.84\%\\
\hline
Densenet161 + Domain  & 86.61\%\\
\hline
Resnet152 + Domain  & 86.49\%\\
\hline
\end{tabular}
\end{center}
\caption{Watermark prediction accuracy on the test data set when the models are retrained end-to-end. Last four rows show the performance after combining domain information}
\label{table:finetuning}
\end{table}

\subsection{Domain based watermark signal}
\label{subsec:domain}
For the images in an image search index, the domain where the images come from is also a very strong signal. Many watermarked images in the web index are coming from stock photo websites. The deep CNN based classifier can not achieve 100\% prediction accuracy on these images. However, a domain based watermark classifier can achieve a higher precision on predicting watermarked images coming from these websites.

In the training data, we group images based on the domains where those images are hosted. We compute the percentage of watermarked images in each domain, which is the ratio of the number of the watermarked images to the all images hosted on this domain. We select domains which produce more than 5 images and have a watermark rate higher than 90\%. In the training data set, there are about 4.7K domains out of about 272K domains that satisfy this condition. We put these domains in a known watermark domain list. For any image coming from those domain, we will predict that this image has a visible watermark regardless of the prediction of the deep CNN classifier. \textbf{Table~\ref{table:domainlist}} shows a few domains containing high percentage watermarked images.

The downside of the domain based approach is that we must have the domain information of the image source. This is not biologically plausible as humans do not need other information besides the raw image to detect the watermark. Also, the domain is dynamic information that can change over time. However, this information is common in images collected from the web. When using this domain information together with the content based watermark information, the accuracy on the validation can be improved as shown in the last four rows of the \textbf{Table~\ref{table:finetuning}}.
\begin{table}
\begin{center}
\begin{tabular}{|l|l|}
\hline
\multicolumn{2}{|c|}{Watermark Domain List}\\
\hline\hline
1. clipartartists.com & 2. www.gettyimages.com\\
\hline
3. www.alamy.com  & 4. www.shutterstock.com\\
\hline
5. www.dreamstime.com & 6. www.cosplayfancy.com\\
\hline
7. www.teamclipart.com  & 8. www.colourbox.de\\
\hline
9. www.recipestable.com  & 10. www.sheepskintown.com\\
\hline
\end{tabular}
\end{center}
\caption{A sample list of a few domains which contain many watermarked images}
\label{table:domainlist}
\end{table}

\section{The Metric}
\label{sec:metric}
\subsection{LambdaMART Ranking Algorithm}
LambdaMART~\cite{lambdamart} is a widely used algorithm in information retrieval to train image ranker. It is built on MART~\cite{mart}. MART builds a regression tree to model the functional gradient of the cost function of interest which leads to the LambdaRank~\cite{lambdarank} functional gradients. For more details, we refer to the corresponding literatures~\cite{mart, ranknet,lambdarank, lambdamart}.

In information retrieval, the widely used metric is the normalized discount cumulative gain (NDCG).  During ranker training, each document has a list of features and an associated rating. The LambdaMart model uses these features and rating of the document to optimize the metric and produce a predicted rank score by which the documents are finally ranked.

LambdaRank can be applied to any image relevane (IR) metric. In the original LambdaRank \cite{lambdarank} paper, the NDCG is defined as
\begin{IEEEeqnarray}{rCl}
\label{eq:NDCG}
N_i = n_i\sum_{j=1}^{T}(2^{r(j)} - 1)/log(1 + j)
\end{IEEEeqnarray}
where $r(j) \in \{0, ..., 4\}$ is the integer label for the relevance level of $j^{th}$ URL in the sorted list. $n_i$ is the normalization factor. However, the rating $(2^{r(j)} - 1)$ only considers image relevance. As a result, the ranker will not be able to pick up the watermark features even when they are available. In our application, instead of using a pure relevance rating, we have a mixed rating score for each image. The rating combined the relevance, image attractiveness and watermark via metric learning. For the image having watermarks, the attractiveness rating will be multiplied by a penalty factor. 

In the following subsections, we will discuss how we get the labels and train the hybrid metric based on these labels.
\subsection{Relevance and Watermark label}
The relevance and watermark labeling is relatively straightforward. For the relevance label, if the query perfectly matches the image content, the image will be labeled as 'Exellent'. If the image content matches the main content of the query, it is labeled as 'Good'. If the image does not cover the main content of the query, it is labeled as 'Bad'. For watermark, if the image contains a watermark, it is labeld as `1', otherwise `0'.

\subsection{Image attractiveness labeling}
\begin{figure}
\begin{center}
\includegraphics[width=0.9\columnwidth]{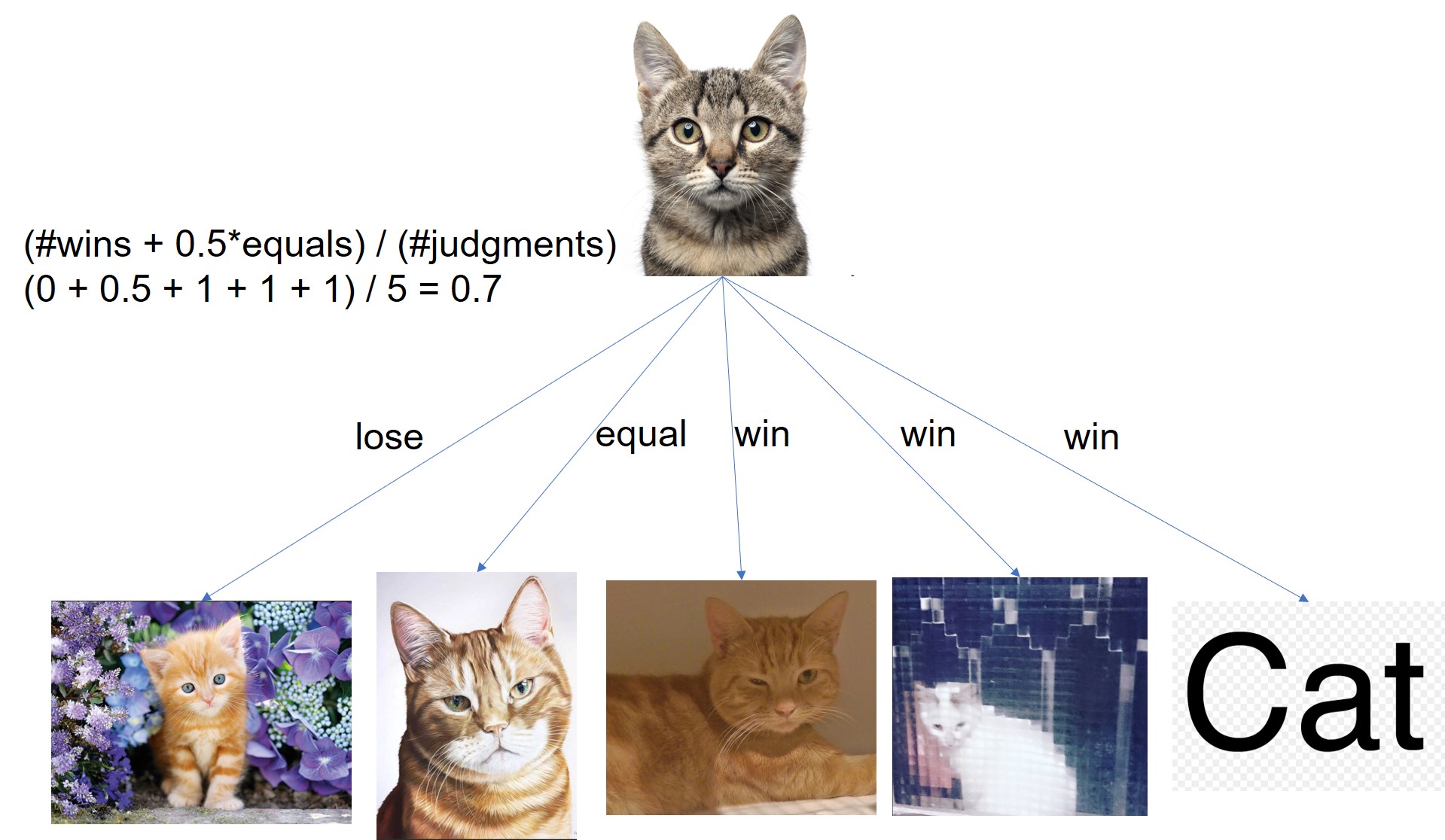}
\end{center}
   \caption{In each query, the image is compared against at most five reference images. The judge will rate if the image is better than the reference images. The judged attractiveness score is computed by $(n\_win + 0.5*n\_equal)/n\_judgments$.}
\label{fig:attlabel}
\end{figure}

\begin{figure}
\begin{center}
\includegraphics[width=0.9\columnwidth]{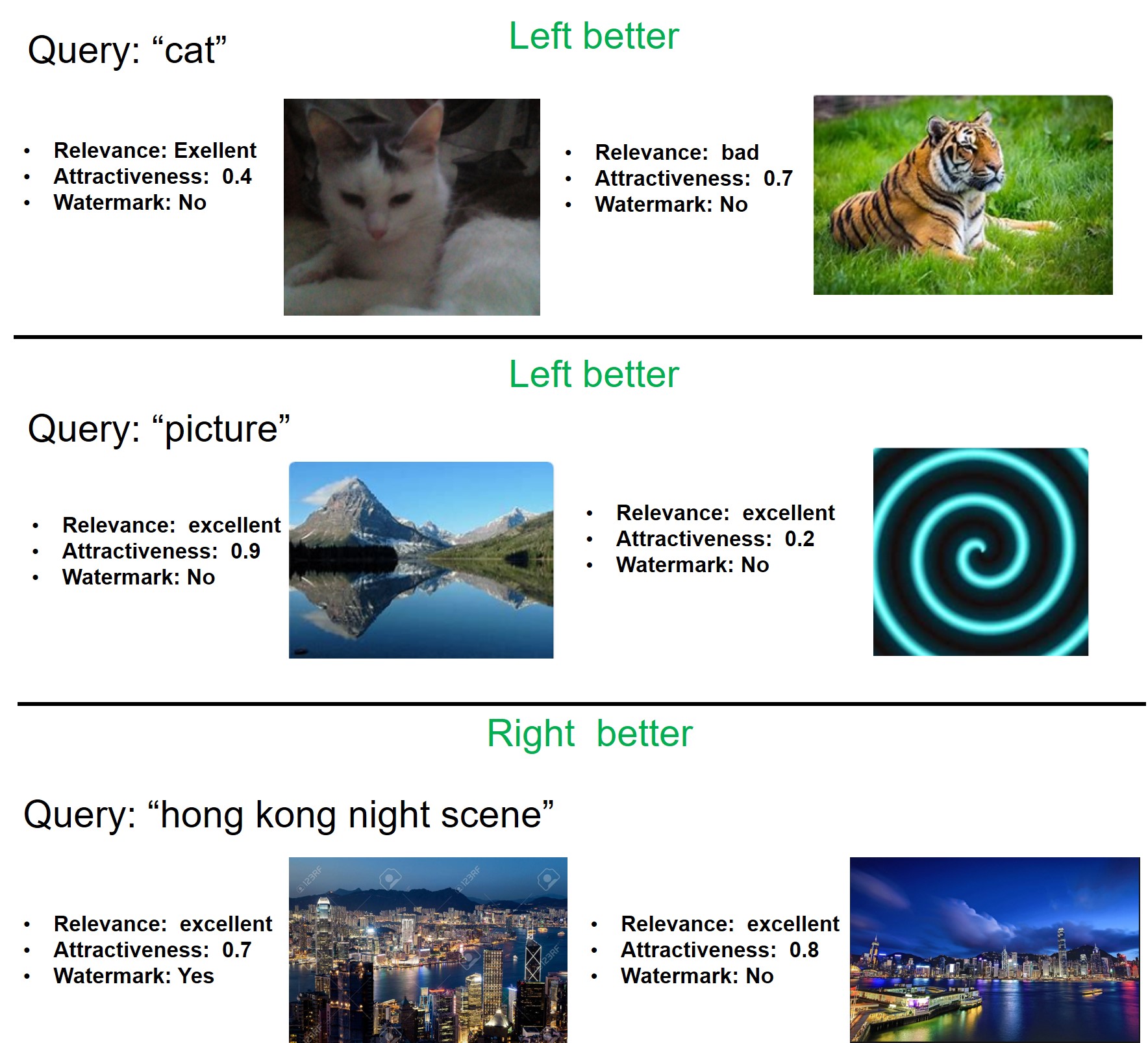}
\end{center}
   \caption{For a specific query, two images are compared side-by-side and rated by five judges with five labels - left better, left light better, equal, right slight better and right better. The judge makes the final decision based on the overall assessment of the image relevance to the query, image attractiveness and the watermark.}
\label{fig:sbsmetriclabel}
\end{figure}

\label{sec:metric_graph}
\begin{figure}
\begin{center}
\includegraphics[width=0.9\columnwidth]{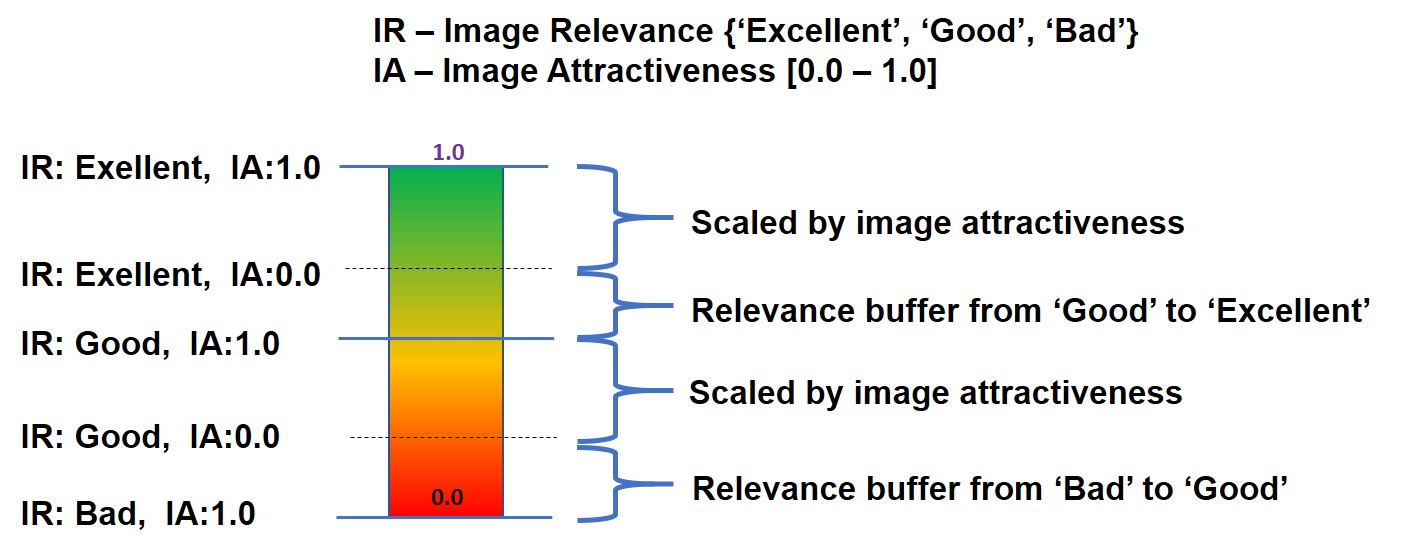}
\end{center}
   \caption{The metric shows how the rating changes as the label changes. For example, an image with 'Excellent' relevance but 0 IA score still has rating larger than an image with 'Good' relevance label and 1 IA score by a certain margin.}
\label{fig:metricgraph}
\end{figure}

For a specific query, we scrap 30 images for this query and compute their corresponding image attractiveness score using DARN \cite{attractivenessning} model. We select at most five representive images based on the attractiveness according to their ranking percentile 100\%,75\%, 50\%, 25\%, 0\%. We call this five images as the reference images set for this query. For each image to be judged, we let the judge compare the image against the five reference images by rating it with `win', `loss' or `equal'. We compute the judged attractiveness score as $\mathrm{(n\_win + 0.5*n\_equal)/n\_judgments}$ where $\mathrm{n\_win}$, $\mathrm{n\_equal}$ and $\mathrm{n\_judgments}$ are the number of wins, loss and total judgments, respectively. The Figure~\ref{fig:attlabel} shows how the judged attractiveness score is computed for a 'cat' image.

\subsection{Side by side labeling}
\label{subsec:sbslabel}
\begin{figure*}
\begin{center}
\includegraphics[width=2\columnwidth]{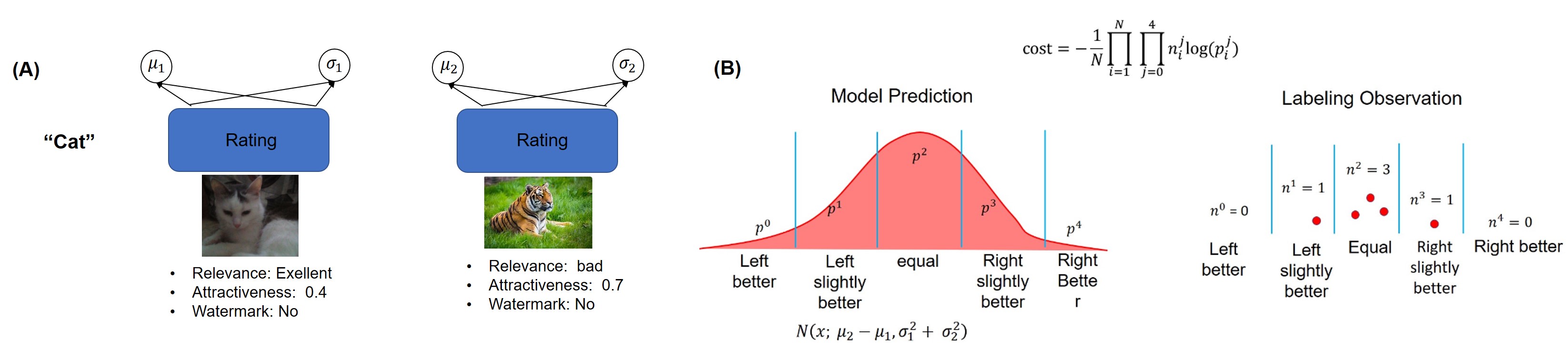}
\end{center}
   \caption{The DARN learns a mapping from an image to its score mean and variance, and a mapping from score difference to the human label. \textbf{Left column}: The distribution of the score difference between images in one pair and the corresponding human pairwise labeling observation. The boundaries are used to map each pair to a label according to their score difference.\textbf{Right column}: The deep attractiveness rank net. The high level features of each image is extracted by a deep CNN. A pairwise DNN is used to map these features to a score mean and variance. The decision boundaries are learned simultaneously.}
\label{fig:loss}
\end{figure*} 

The remaining problem is how do we design a hybrid new rating which combines the relevance, image attractiveness and watermark together in one place. The rating needs to solve a few questions - (1) how do we weight the importance between relevance, image attractiveness, and watermark; (2) how the watermark will effect the rating; and (3) how the model can have the freedom to learn that the relevance is the most important factor which is usually required in image retrieval.

To learn the hybrid rating, we select 200 queries and scrape 30 images for each query. The data used to train the metric is independent of the data we used to train the watermark classifier and image ranker. Within each query, we randomly pairs the images. Each image in a pair will have relevance label, image attractiveness label, and watermark label provided by the judges. Out of these images, about 10\% contains watermarks. Additionally, we let judge decide which image is better overall for this particular query, by choose five labels "left better, left slight better, equal, right slight, better". According to this relative pairwise labeling, we can directly learn the metric weights among the three factors. Figure \ref{fig:sbsmetriclabel} demonstrates the process of this side-by-side labeling procedure. For example, in the first pair, the judge chooses that the left image is better even its image attractiveness is poor. The rating learns that image relevance is more important. In the second pair, the judge chooses the right image better because it is more appealing. The rating learns that image attractiveness is important. In the third pair, the judge thinks right is better as the left image has watermark and is less attractive. So, the rating will learn the importance of watermark and image attractiveness. Eventually, the rating should be able to learn how important each factor is and finally converges to a hybrid rating reflecting a user's overall experience.

Eq~(\ref{eq:IR}) to Eq~(\ref{eq:rate}) is our design of the new hybrid rating. Each image's rating depends on three labels - the relevance label $\mathrm{\mathbf{IR}}$:\{0:Bad, 1:Good, 2:Exellent\}, the attractiveness label $\mathrm{\mathbf{IA}}\in[0,1]$, and the watermark label $\mathrm{\mathbf{WM}}$:\{0:No Watermark, 1:Has visible watermark\}. $\mathbf{RatingIRs}= \mathbf{[}\mathbf{0}, \mathbf{RatingGood},\mathbf{RatingExellent}\mathbf{]}$ is a learnable parameter vector
where $\mathrm{\mathbf{RatingIRs[IR]}}$ denotes the best rating an image will get for this relevance label $\mathbf{IR}$. Here, `best rating' is the rating an image will get which has attractiveness score 1.0 and no watermark. $\mathrm{\mathbf{Gamma}}$ is also a learnable parameter which defines the rating buffer between two consecutive relevance label. It is to ensure that the rating of an image with better relevance label is larger than the one of another image with worse relevance label by a certain margin. $\mathbf{WMP}$ stands for watermark penalty which denotes additional penalty it will apply on the images attractiveness score for an image containing watermark. For a given relevance label $\mathbf{IR}$, in Eq~(\ref{eq:IR}) and Eq~(\ref{eq:IRPrev}), we get the rating for this relevance label $\mathbf{RatingIRs[IR]}$, and the rating for the relevance downgraded by one level $\mathbf{RatingIRs[IR - 1]}$. The score is computed via Eq~\eqref{eq:rate} to Eq~\eqref{eq:rate2}. Eq~\eqref{eq:rate1} is the buffer region between two relevance labels. Intuitively, it means that if an image gets a relevance label 'Exellent' even with 0 image attractiveness score and containing watermark, this image's rating will still be more than another image, which has only 'Good' relevance label but perfect secondary score, by the margin of $\mathbf{\mathbf{BucketWidth} * (1 - \mathbf{Gamma})}$. This gives the model the freedom to learn that the relevance is always the dominate factor. Then, the image score will be further scaled by the image attractiveness as shown in Eq~\eqref{eq:rate2}.

\begin{IEEEeqnarray}{rCl}
\label{eq:IR}
\mathrm{RatingIR} & = &\mathrm{RatingIRs}[\mathrm{IR}]
\end{IEEEeqnarray}
\begin{IEEEeqnarray}{rCl}
\label{eq:IRPrev}
\mathrm{RatingIRPrev} & = &\mathrm{RatingIR}[\mathrm{max}(\mathrm{IR} - 1,0)]
\end{IEEEeqnarray}
\begin{IEEEeqnarray}{rCl}
\label{eq:Bucket} 
\mathrm{BucketWidth} & = &\mathrm{RatingIR} - \mathrm{RatingIRPrev}
\end{IEEEeqnarray}
\begin{IEEEeqnarray}{rCl}
\label{eq:IA}
\mathrm{IA} & = &(1 - \mathrm{WMP} * \mathrm{WM}) * \mathrm{IA}
\end{IEEEeqnarray}
\begin{IEEEeqnarray}{rCl}
\IEEEyesnumber\label{eq:rate} \IEEEyessubnumber*
\mathrm{Rating} & = &\mathrm{RatingIRPrev}\\ \label{eq:rate1}
& + &  \mathrm{BucketWidth} * (1 - \mathrm{Gamma}) \\ \label{eq:rate2}
& + & \mathrm{IA} * \mathrm{BucketWidth} * \mathrm{Gamma} \label{eq:sub3}
\end{IEEEeqnarray}
\begin{IEEEeqnarray}{rCl}
\label{eq:rating}
\mathrm{Rating} & = &\mathrm{Rating} / \mathrm{max(RatingIR)}
\end{IEEEeqnarray}

\begin{figure*}
\begin{center}
\includegraphics[width=2\columnwidth]{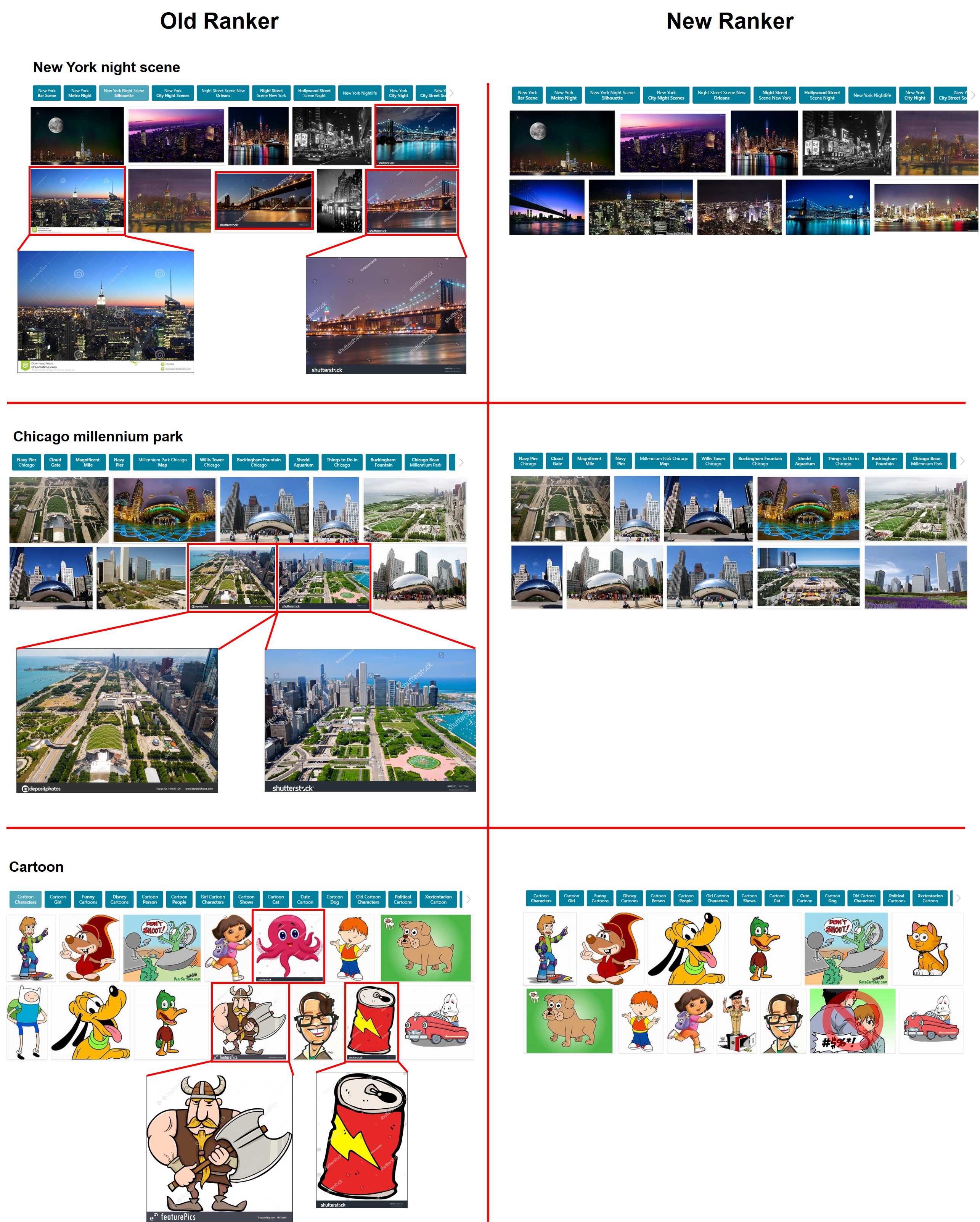}
\end{center}
   \caption{The watermarked images, denoted by red crosses, are demoted in the ranker using watermark signal (right) compared to the control ranker where no watermark signal is used (left). In order to see the details of the watermarks, two selected watermarked images are zoomed out in each case. For example, in the first row, the old ranker surfed 4 watermarked images while the new ranker surfed none.}
\label{fig:rankercompare}
\end{figure*}

\subsection{Metric Loss function}
We use similar pairwise rank loss proposed in \cite{attractivenessning} to learn the rating. We would like to learn the rating with pairs of images $[x_1,x_2]$ judged side-by-side with a relative label $Y$ as demonstrated in \ref{subsec:sbslabel}. The goal is to learn a model $f:\mathbb{R}^d \mapsto \mathbb{R}$ such that the images with higher rating  (i.e., $f(x_1) > f(x_2)$ indicates image $x_1$ is better than $x_2$ when assessed based on the labels. Figure~\ref{fig:loss} graphically illustrates metric learning structure. The model is designed to predict the mean and variance of the rating for an image as shown in Figure~\ref{fig:loss}A, and the decision boundary that specifies how differences in these distributions correspond with judge preferences as shown in Figure~\ref{fig:loss}B.

\begin{figure}
\begin{center}
\includegraphics[width=0.9\columnwidth]{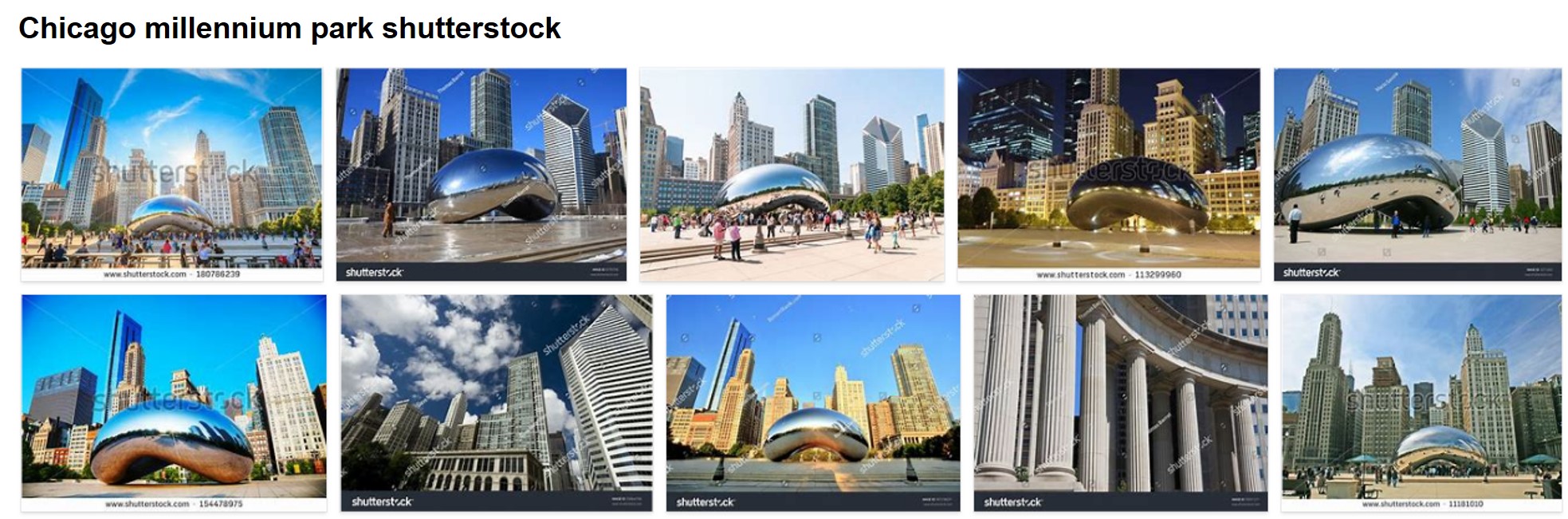}
\end{center}
   \caption{This shows that if the query explicitly asks for images from some websites, such as 'shutterstock', which contains visible watermarks, the search engine can still return the images from 'shutterstock' because of the relevance.}
\label{fig:shutter}
\end{figure}

Let us define $\mu = \mathrm{E}[f(x)]$ and $\sigma^2 = \mathrm{Var}[f(x)]$ as the mean rating score generated by Eq~(\ref{eq:rating}) and variance for an image, respectively. For an image pair $[\mathbf{x_i}:(\mu_i, \sigma_i), \ \mathbf{x_j}:(\mu_j, \sigma_j)]$, define $\mathrm{P_{ij}}(\mathbf{x_i}\leftrightarrow\mathbf{x_j} = y)$ as the posterior probability that the image pair is labeled as $y$. 

Assume each image can be rated by a large number of experts who have extremely high confidence of the overall rating.  According to central limit theorem, the rating scores received by each image will follows a normal distribution $\mathcal{N}(x;\mu,\sigma)$. Again, the $\mu$ denotes the rating in Eq~(\ref{eq:rating}) and $\sigma$ stands for variance. For the two images $x_i$ and $x_j$, the score difference is also a normal distribution $\mathcal{N}(x;\mu_i - \mu_j,\sigma_i^2 + \sigma_j^2)$ as shown in left panel of \textbf{Figure~\ref{fig:loss}(A)}.
The model learns four boundaries which are used to map each pair to a label according to their score difference. We define the four boundaries as $\{b_i\}_{i = 0}^3$. Let $p_i^j$ denote the probability that the $i_{th}$ pair is labeled as $j$ (indexing \{left better, left slightly, equal, right sightly, right better\} as \{0, 1, 2, 3, 4\}), and $\bigtriangleup\mu_i$ and $\bigtriangleup\sigma_i$ as the mean and variance of the score difference of $i_{th}$ pair. $p_i^j$ is the probability of the $i_{th}$ pair labeled as $j$ and is represented by the area under the normal distribution of the score difference of the $i_{th}$ pair between boundary $b_{j - 1}$ and $b_j$, i.e., $p_i^j = \int\limits_{b_{j - 1}}^{b_{j}}\mathcal{N}(x;\bigtriangleup\mu_i, \bigtriangleup\sigma_i)dx$.  Let $n_i^j$ indicate the number of judges labeling $i_{th}$ pair as label $j$. Thus, we define a log maximum likelihood cost function as 

\begin{IEEEeqnarray}{rCl}
cost  = -\log \prod\limits_{i = 1}^{N}\prod\limits_{j = 0}^{4}(p_i^j)^{n_i^j} = -\sum\limits_{i = 1}^{N}\sum\limits_{j = 0}^{4}n_i^j\log (p_i^j)
\end{IEEEeqnarray}
where N is the number of pairs.

We use backpropagation to jointly learn the decision boundaries, the variance, and the parameters to compute the hybrid rating in Eq~(\ref{eq:IR}) to Eq~(\ref{eq:rating}). Then, each image document will get an overall rating using this learned rating based on their labels. This new rating replaces the original rating in Eq (\ref{eq:NDCG}) and is used during image ranker training.

\section{Online results of utilizing watermark signal}
\label{sec:onlinerank}
During ranker training, each document's rating is computed using the learned new hybrid rating. The image ranker will take a list of documents each of which associates with a new rating and a list of features. LambdarMart algorithm ia used to trained the image ranker. We trained two image rankers. The control ranker uses the original features during training. In the experimental ranker training, we add the watermark signal obtained from deep CNN model and domain analysis into the existing feature pool. If the image's doamin does not belong to the domain black list, we will use the watermark probability predicted by the Resnet50 model. Otherwise, the watermark probability is 1. \textbf{Table~\ref{table:watermarkrate}} shows that the experimental ranker's watermark rate is reduced from 5.2\% to 4.7\%, relatively by 10\% after adding the domain based watermark feature. The watermark rate is futher decreaded to 3.7\%, relatively by 20\% after adding the DNN based watermark feature. The NDCG is also improved. \textbf{Figure~\ref{fig:rankercompare}} shows a few examples where the watermarked images are demoted for a few example queries. For example,  for the query `New York night scene' in the first row, the control ranker surfaced 5 watermarked images (highlighted by red bounding box),
while the new ranker shows no watermarked images.

\begin{table}
\begin{center}
\begin{tabular}{|l|c|c|}
\hline
Ranker & Watermark Rate & NDCG\\
\hline\hline
No watermark signal & 5.2\% & 62.3\\
\hline
Domain watermark signal & 4.7\% & 62.4\\
\hline
DNN watermark signal & 3.9\% & 62.5\\
\hline
Both watermark signals & 3.8\% & 62.5\\
\hline
\end{tabular}
\end{center}
\caption{Using watermark signal in the ranker significantly decreased the online watermark rate and increased NDCG.}
\label{table:watermarkrate}
\end{table}

\section{Discussion}
We proposed a few techniques to obtain watermark signals from online images and demonstrated the effectiveness of utilizing them in the image search. We designed a hybrid metric for the image ranker to enable it to pick up watermark related features. This sheds light on the solution to provide better image search quality to the user by effectively demoting watermarked images. More research can also be done to understand what image attributes are mainly responsible for watermark detection and which part of the neural network is sensible for the watermark information.
\section{Acknowledgement}
We thank Rui Xia, Viktor Burdeinyi, Yiran Shen, Houdong Hu and Arun Sacheti for valuable help.

\clearpage
{\small
\bibliographystyle{ieee}
\bibliography{egbib}

\begin{thebibliography}{10}\itemsep=-1pt

\bibitem{lambdarank}
C.~Burges, R.~Ragno, and Q.~Le.
\newblock Learning to rank with non-smooth cost functions.
\newblock In {\em Conference on Neural Information Processing Systems (NIPS)},
  2006.

\bibitem{ranknet}
C.~Burges, T.~Shaked, E.~Renshaw, A.~Lazier, M.~Deeds, N.~Hamilton, and
  G.~Hullender.
\newblock Learning to rank using gradient descent.
\newblock In {\em International Conference on Machine Learning (ICML)}, 2005.

\bibitem{removal_Dashti}
M.~Dashti, R.~Safabakhsh, M.~Pourfard, and M.~Abdollahifard.
\newblock Video logo removal using iterative subsequent matching.
\newblock In {\em The International Symposium on Artificial Intelligence and
  Signal Processing (AISP)}, 2015.

\bibitem{WatermarkremovalGoogle}
T.~Dekel, M.~Rubinstein, C.~Liu, and W.~Freeman.
\newblock On the effectiveness of visible watermark.
\newblock In {\em Conference on Computer Vision and Pattern Recognition
  (CVPR)}, 2017.

\bibitem{ImageNet}
J.~Deng, W.~Dong, R.~Socher, L.~Li, K.~Li, and F.~Li.
\newblock Imagenet: A large-scale hierarchical image database.
\newblock In {\em IEEE Conference on Computer Vision and Pattern Recognition
  (CVPR)}, 2009.

\bibitem{mart}
J.~Friedman.
\newblock Greedy function approximation: A gradient boosting machine.
\newblock In {\em Technical report, Dept. Statistics, Stanford, 1999}, 1999.

\bibitem{Resnet}
K.~He, X.~Zhang, S.~Ren, and J.~Sun.
\newblock Deep residual learning for image recognition.
\newblock In {\em Conference on Computer Vision and Pattern Recognition
  (CVPR)}, 2016.

\bibitem{Densenet}
G.~Huang, Z.~Liu, L.~Maaten, and K.~Weinberger.
\newblock Densely connected convolutional networks.
\newblock In {\em Conference on Computer Vision and Pattern Recognition
  (CVPR)}, 2017.

\bibitem{attractivenessning}
N.~Ma, A.~Volkov, A.~Livshits, P.~Pietrusinski, H.~Hu, and M.~Bolin.
\newblock An universal image attractiveness ranking framework.
\newblock In {\em IEEE Winter Conference on Applications of Computer Vision
  (WACV)}, 2019.

\bibitem{InceptionV3}
C.~Szegedy, V.~Vanhoucke, S.~Ioffe, J.~Shlens, and Z.~Wojna.
\newblock Rethinking the inception architecture for computer vision.
\newblock In {\em Conference on Computer Vision and Pattern Recognition
  (CVPR)}, 2016.

\bibitem{Talebi2017}
H.~Talebi and P.~Milanfar.
\newblock Nima: Neural image assessment.
\newblock {\em IEEE Transactions on Image Processing}, 27:3998--4011, 2017.

\bibitem{removal_Wang}
J.~Wang, Q.~Liu, L.~Duan, H.~Lu, and C.~Xu.
\newblock Automatic tv logo detection, tracking and removal in broadcast video.
\newblock In {\em 13th International Multimedia Modeling Conference (MMM)},
  2007.

\bibitem{lambdamart}
Q.~Wu, C.~Burges, K.~Svore, and J.~Gao.
\newblock Ranking, boosting, and model adaptation.
\newblock In {\em Microsoft Research Technical Report MSR-TR-2008-109}, 2008.

\bibitem{removal_Yan}
W.~Yan, J.~Wang, and M.~Kankanhalli.
\newblock Automatic video logo detection and removal.
\newblock In {\em Multimedia Systems}, 2005.

\end{thebibliography}
}
\end{document}